\documentclass{article}
\usepackage{iclr2027_conference,times}

\usepackage{amsmath,amsfonts,bm}

\def\eqref#1{equation~\ref{#1}}
\def\1{\bm{1}}

\DeclareMathAlphabet{\mathsfit}{\encodingdefault}{\sfdefault}{m}{sl}
\SetMathAlphabet{\mathsfit}{bold}{\encodingdefault}{\sfdefault}{bx}{n}

\usepackage{hyperref}
\usepackage{url}

\usepackage{graphicx}
\usepackage{booktabs}
\usepackage{multirow}
\usepackage{makecell}
\usepackage{xspace}
\usepackage{xcolor}
\usepackage{colortbl}
\usepackage{siunitx}
\usepackage{verbatim}
\usepackage{cleveref}

\definecolor{sgreen}{HTML}{F3FADF}  % lightest
\definecolor{mgreen}{HTML}{E0EAB5}  % medium-light
\definecolor{dgreen}{HTML}{CDDC8C}  % medium-dark
\definecolor{ddgreen}{HTML}{B8CF61} % darkest

\newcommand{\method}{\textsc{SaeRL}\xspace}
\newcommand{\methodsel}{\textsc{SaeRL}\textsubscript{Sel}\xspace}
\newcommand{\methodcur}{\textsc{SaeRL}\textsubscript{Curr}\xspace}
\newcommand{\methodbatch}{\textsc{SaeRL}\textsubscript{Batch}\xspace}
\newcommand{\methodord}{\textsc{SaeRL}\textsubscript{Ord}\xspace}

\newcommand{\methodd}{\textsc{SaeRL}\textsuperscript{DAPO}\xspace}

\title{
Teach the Way It Thinks: \\
Guiding LLM Post-training Data Engineering with Sparse Autoencoders
}

\author{{\bfseries Yi Jing\thanks{Equal contribution.}\quad Zao Dai\footnotemark[1]\quad Jinwu Hu\quad Zijun Yao\quad Lei Hou\quad Juanzi Li\quad Xiaozhi Wang} \\
  Tsinghua University \\
  {\small\texttt{\{jingy22,dz26\}@mails.tsinghua.edu.cn}}}

\iclrfinalcopy

\begin{document}

\maketitle

\begin{abstract}
Data engineering for large language model (LLM) reinforcement learning (RL) decides which samples to train on, when to present them, and which to batch together. 
Existing methods mostly rely on external feedback, such as pass rates or difficulty labels, which observes only the outcome on each sample. We argue that model internals offer richer and more model-intrinsic feedback that can guide all three decisions.
We propose \method, which organizes RL training data in the feature space of Sparse Autoencoders (SAEs), a mechanistic interpretability tool that decomposes model internals into sparse, interpretable features. 
SAE features enable viewing the model as a set of functional regions, each formed by features that related samples activate together. For each sample, its SAE activations indicate which regions it engages, and its gradients projected onto SAE features indicate how training on it would change these regions. 
Guided by this view, \method keeps samples with strong and non-redundant gradient signals, trains each functional region from easy to hard, and trains one functional region at a time, drawing each batch from a single activation cluster with moderate cross-cluster mixing. 
The resulting curriculum is constructed once with small models and transfers to larger ones. Experiments on mathematical reasoning show consistent gains across model scales and RL algorithms. For instance, on \texttt{Qwen2.5-Math-1.5B}, \method improves the average accuracy of GRPO by $3.1\%$ and reaches the target accuracy with $42\%$ fewer training steps. 
Further analyses reveal that ordering and batching are complementary. These results demonstrate that model internals are a powerful and practical foundation for post-training data engineering. The source code is available at \url{https://github.com/THUAIS-Lab/SAERL}.
\end{abstract}

\section{Introduction}
% \begin{figure}[!htb]
%     \begin{center}
%     \includegraphics[width=0.78\linewidth]{figures/intro.pdf}
%     % \includegraphics[width=0.98\linewidth, trim=10 150 10 100, clip]{figures/intro.pdf}
%     \end{center}
%     \caption{Conceptual overview of \method. Sparse Autoencoder (SAE) activations characterize three intrinsic data properties (diversity, difficulty, and quality), enabling SAE-based curriculum learning and data selection for LLM post-training.}
%     \label{fig:inrto}
% \end{figure}

Post-training, especially reinforcement learning, has become central to advancing the capabilities of large language models \citep{openai2026gpt6astra,anthropic2026claudeopus55,zeng2026glm,deepseekai2026deepseekv41flash,kimi2026k3}. Its effectiveness depends heavily on data engineering, which answers three questions: which samples to train on, when to present them, and which samples to present together in a batch. These choices shape the training signal at every step, making data engineering an important factor for improving both training efficiency and final performance.

Existing post-training data engineering pipelines typically rely on external feedback signals, including human preferences~\citep{ouyang2022training,lambert2024tulu3}, verifier outcomes~\citep{deepseekai2025deepseekr1,shao2024deepseekmath,yu2025dapo}, rollout pass rates~\citep{sun2025dots,xu2025pods,zheng2025greso}, and difficulty signals~\citep{narvekar2020curriculum,shi2025adarft,gao2025promptcurriculum,zhao2025ufo}.
These signals have proven useful for data selection and curriculum learning.

However, external feedback is costly to obtain during training~\citep{casper2023open} and observes only the outcome on each sample.
A model-intrinsic perspective could provide richer feedback by characterizing which components of the model each sample engages and how training on the sample updates these components.
Prior work approximates such a perspective with raw activations and gradients, which have been shown to guide data selection in pre-training~\citep{sam2025analyzing,rathi2026shaping} and supervised fine-tuning~\citep{ivison2025large,ma2025task,chen2026neuron}.
Yet raw activations and gradients superimpose many concepts along the same dimensions~\citep{elhage2022superposition}, offering only an indirect view of which components a sample engages.

Mechanistic interpretability research~\citep{meng2022locating,wang2022interpretability,somvanshi2026bridging} continuously explores how to disentangle and understand model internals.
As a recent advance, Sparse Autoencoders (SAEs) decompose LLM hidden representations into sparse, fine-grained feature activations~\citep{bricken2023towards,gao2024scaling,
templeton2024scaling}, providing a disentangled view of model internals: individual features respond to specific concepts, and related inputs activate overlapping groups of features, much as related tasks engage the same functional region of the brain. This makes SAE features a natural representation space for data engineering: sample activations can be encoded as sparse feature vectors, and policy gradients can be projected onto SAE feature directions, placing both internal signals on a shared set of fine-grained features.

We therefore propose \method, a data engineering framework that organizes RL post-training data in the SAE feature space. Its design follows three principles, each answering one of the three questions posed earlier:
(1) \emph{Train where the model can still improve}: \method keeps samples with strong gradient signals on SAE features, drops those whose gradient patterns repeat stronger ones, and retains samples that the small model fails to solve.
(2) \emph{Train each functional region from easy to hard}: \method clusters samples by their SAE activations, sorts each cluster by a difficulty score predicted from SAE features, and interleaves clusters over training.
(3) \emph{Train one functional region at a time}: \method draws each batch from a single activation cluster, so that its update concentrates on a shared group of SAE features, and exchanges a few samples across clusters to keep other regions covered.
All steps are performed offline with small models and their SAEs, so the resulting curriculum is computed once and reused for training larger models.

Experiments on mathematical reasoning show that \method consistently improves both performance and training efficiency across model scales and RL algorithms. On \texttt{Qwen2.5-Math-1.5B}, it improves the average accuracy of GRPO by $3.1\%$ and reaches the target accuracy with $42\%$ fewer training steps. Guided by SAEs from small models, \method also improves the average accuracy on \texttt{Qwen2.5-Math-7B} by $1.9\%$ while achieving the fastest convergence among all methods, showing that the curriculum constructed with small models transfers to larger ones.

Our contributions are as follows. (1) We show that model internals can serve as rich, model-intrinsic feedback for post-training data engineering, complementary to external feedback. (2) We propose \method, which selects, orders, and batches RL training data with SAE features, using a curriculum built once on small models and transferable to larger ones. (3) Experiments across model scales and RL algorithms show that \method consistently improves both performance and training efficiency, with analyses of how each operation contributes. We hope that this work can facilitate future research on intrinsic data engineering and actionable mechanistic interpretability~\citep{orgad2026interpretability}. 

\begin{figure}[t]
    \centering
    \begin{minipage}[t]{0.43\linewidth}
        \centering
        \includegraphics[width=\linewidth]{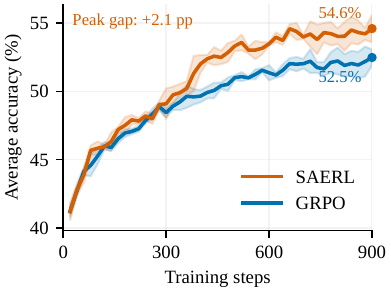}
        \par\textbf{(a) Training performance}
    \end{minipage}
    \hfill
    \begin{minipage}[t]{0.43\linewidth}
        \centering
        \includegraphics[width=\linewidth]{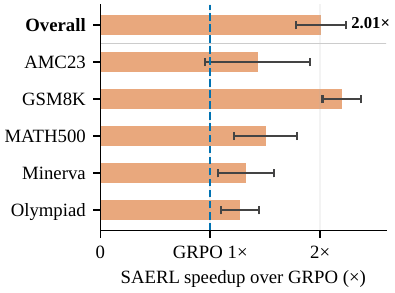}
        \par\textbf{(b) Training efficiency}
    \end{minipage}
    \caption{On \texttt{Qwen2.5-Math-1.5B}, \method achieves higher average
    accuracy and reaches GRPO's peak in fewer steps across five math benchmarks: AMC23, GSM8K, MATH500, MinervaMath, and OlympiadBench. Each method is trained with three different seeds; shading and error bars show standard deviations over three runs.}
    \label{fig:intro}
\end{figure}

\section{Methodology}
\label{sec:sae-rl-pipeline}

\method reads model-intrinsic feedback for each sample through SAE features: its activations indicate which functional regions of the model it engages, and its gradients projected onto SAE features indicate how training on it would update these regions. We first formalize the problem and outline the pipeline (\Cref{subsec:overview}), and then detail feature extraction, data selection, and curriculum construction (\Cref{subsec:sae-repr,subsec:selection,subsec:curriculum}).

\subsection{Overview}
\label{subsec:overview}

\begin{figure}[!ht]
    \centering
    \includegraphics[width=0.98\linewidth]{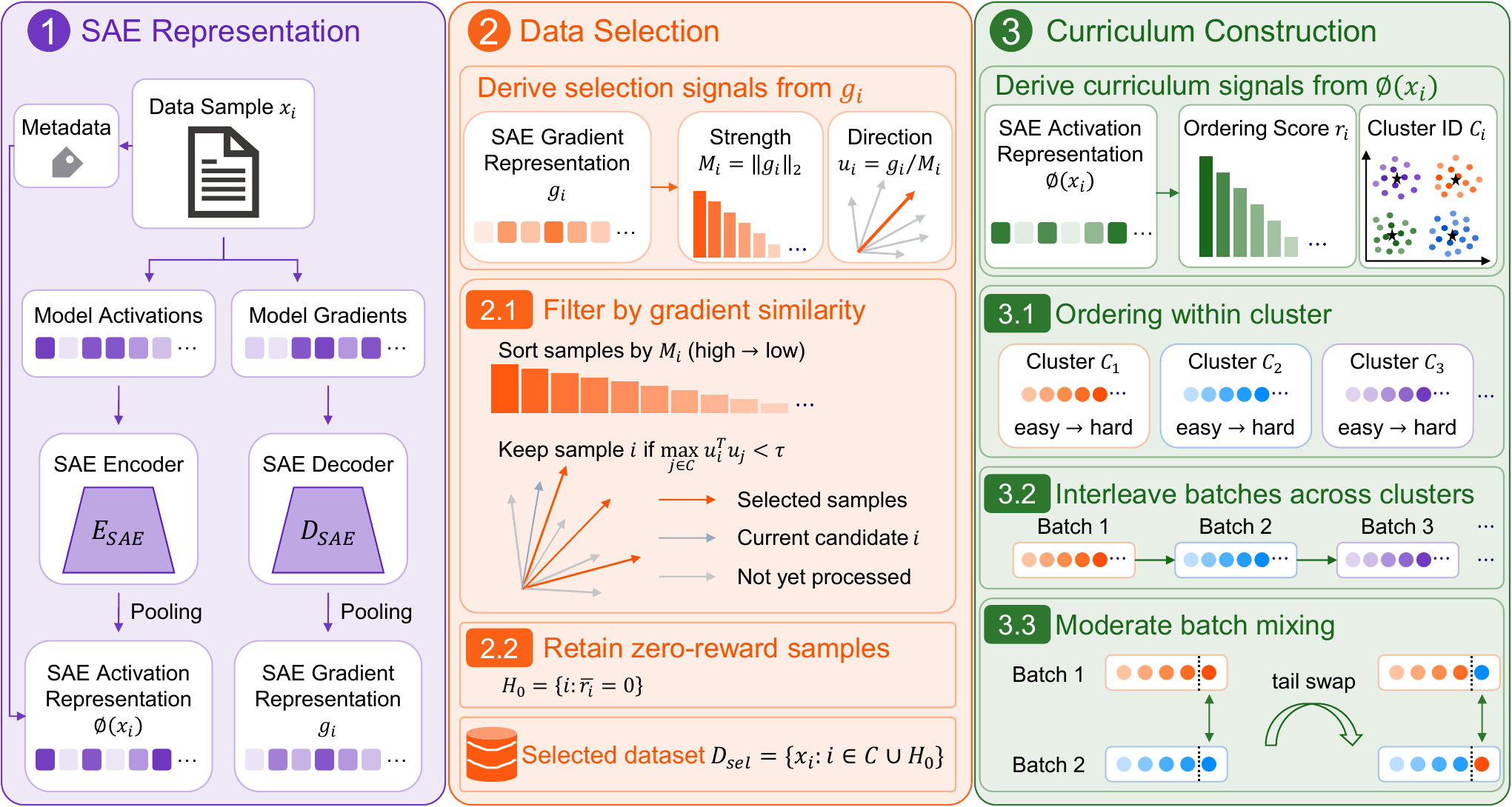}
    \caption{Overview of \method. SAE representations of each sample drive data selection by gradient strength and similarity, and curriculum construction by clustering, easy-to-hard ordering, and moderate batch mixing.}
    \label{fig:method}
\end{figure}

\paragraph{Problem setup.}
Let $\mathcal{D}=\{x_i\}_{i=1}^{N}$ be the training pool for RL post-training. Given a batch size $b$, data engineering produces an ordered sequence of batches $\mathcal{S}=(B_1,\dots,B_T)$ with $B_t\subseteq\mathcal{D}_{\mathrm{sel}}$ and $|B_t|=b$, which the RL trainer consumes in order. Producing $\mathcal{S}$ involves two operations. \emph{Data selection} chooses a subset $\mathcal{D}_{\mathrm{sel}}\subseteq\mathcal{D}$. \emph{Curriculum construction} partitions $\mathcal{D}_{\mathrm{sel}}$ into batches and orders them, determining when each sample is seen and which samples are seen together.

\paragraph{Framework.}
As shown in \Cref{fig:method}, \method proceeds in three offline steps.
(1) \emph{Feature extraction} (\Cref{subsec:sae-repr}): an SAE trained on a small model encodes each sample into an activation representation, and the small model's rollout gradients are projected onto SAE decoder directions to form a gradient representation.
(2) \emph{Data selection} (\Cref{subsec:selection}): samples with strong and non-redundant gradient signals are kept, together with samples that receive zero reward under the small model, yielding $\mathcal{D}_{\mathrm{sel}}$.
(3) \emph{Curriculum construction} (\Cref{subsec:curriculum}): $\mathcal{D}_{\mathrm{sel}}$ is clustered into functional regions in the activation space and sorted from easy to hard within each region; batches are then interleaved across clusters and moderately mixed, yielding $\mathcal{S}$.
Because all steps rely only on small models and their SAEs, $\mathcal{S}$ is computed once and reused across training runs and model scales.

\subsection{Feature Extraction}
\label{subsec:sae-repr}

Both operations in \method act on sample representations built from SAE features. An SAE decomposes a model's dense hidden activations into sparse feature activations~\citep{bricken2023towards,gao2024scaling}, providing coordinates in which samples can be compared, grouped, and ordered. We construct two representations, each computed with a frozen SAE trained on a small language model.

\paragraph{Activation representation.}
Given a sample $x_i$, we extract token-level SAE activations separately from its prompt and solution spans, and aggregate each span with mean and max pooling to capture both sustained and localized activation patterns. Concatenating the pooled activations over both spans gives $z_i$ ($960$ dimensions), and appending shallow metadata features $m_i$ such as length statistics, TeX ratio, and digit ratio ($26$ dimensions) yields the activation representation $\phi(x_i)=[z_i,m_i]$, which is used for curriculum construction.

\paragraph{Gradient representation.}
To characterize how each sample would update the model, we generate rollouts for $x_i$ with the small model, project token-level response gradients onto SAE decoder directions, and aggregate the projections over tokens and rollouts~\citep{shu-etal-2025-beyond}. The resulting gradient representation $g_i$ describes how the sample's gradient is distributed across SAE features and is used for data selection. Details of both representations are provided in Appendix~\ref{app:sae-representation}.

\subsection{Data Selection}
\label{subsec:selection}

Data selection aims to retain samples that provide strong and non-redundant training signals. Since this concerns how each sample would update the model, we perform selection on the gradient representations $g_i$.

\paragraph{Filtering by gradient similarity.} We use the norm of $g_i$ to measure each sample's signal strength, and cosine similarity to determine whether two samples induce similar gradient patterns across SAE features. 
For samples with nonzero representations, define
\begin{equation}
M_i=\lVert g_i\rVert_2,\qquad
u_i=g_i/M_i.
\end{equation}
We process samples in descending order of $M_i$ so that stronger signals are considered first. Starting from an empty core set $\mathcal C$, a sample is retained only when its similarity to every selected sample is below a threshold $\tau$:
\begin{equation}
\mathcal C\leftarrow\mathcal C\cup\{i\}
\quad\text{if}\quad
\mathcal C=\emptyset
\quad\text{or}\quad
\max_{j\in\mathcal C}u_i^\top u_j<\tau.
\end{equation}
A sample is therefore omitted when a selected sample already provides a similar gradient pattern with an equal or stronger signal; $\tau$ controls how similar two samples must be for one to be omitted.

\paragraph{Retaining zero-reward samples.} Under GRPO, a sample whose rollouts all receive the same reward has zero group-normalized advantages and hence a zero gradient representation, so it never enters the core set. This includes the hard samples that the small model always fails, which the policy should eventually learn. We therefore add all samples with zero average reward $\bar r_i$, computed from the same rollouts used for feature extraction:
\begin{equation}
\mathcal H_0=\{i:\bar r_i=0\},\qquad
\mathcal D_{\mathrm{sel}}=\{x_i:i\in\mathcal C\cup\mathcal H_0\}.
\end{equation}
The resulting $\mathcal D_{\mathrm{sel}}$ is passed to curriculum construction.

\subsection{Curriculum Construction}
\label{subsec:curriculum}

Curriculum construction arranges $\mathcal D_{\mathrm{sel}}$ into the batch sequence $\mathcal S$ and involves two decisions: \emph{ordering}, which determines when each sample appears, and \emph{batching}, which determines which samples share a batch. Both decisions are made on the activation representation $\phi(x_i)$ and share a common clustering. We cluster samples by $\phi(x_i)$ using MiniBatchKMeans~\citep{sculley2010web} with $K{=}10$ and denote the cluster of $x_i$ by $c_i$. We treat each cluster as a functional region of the model: its samples activate overlapping groups of SAE features and thus engage similar components of the model. The resulting clusters do not simply reproduce the dataset-provided topic labels; they reflect combinations of topic, problem format, and solution profile (Appendix~\ref{app:interpretability}).

\paragraph{Ordering.}
We train a lightweight ElasticNet regressor~\citep{zou2005regularization} on a small difficulty-labeled subset ($|\mathcal L|{=}3\text{k}$), producing a raw score $\hat d_i=f_D(\phi(x_i))$ for every sample. Because raw scores may vary in scale across clusters, we calibrate them with a global mapping and shrinkage-based cluster corrections:
\begin{equation}
    s_i = \text{Calibrate}\bigl(\hat{d}_i,\, c_i\bigr),
\end{equation}
where $s_i$ is the calibrated difficulty score (Appendix~\ref{app:difficulty-clustering-calibration}). Within each cluster, samples are sorted by $s_i$ in ascending order, forming an easy-to-hard trajectory for each functional region. These per-cluster sequences are then interleaved into a single global order, so that overall difficulty rises over training while every stage of training draws on multiple clusters.

\paragraph{Batching.}
Each cluster's sorted sequence is partitioned into batches of size $b$, so each batch initially focuses on a single functional region with samples of similar difficulty, concentrating its update on a shared group of SAE features. To keep other regions covered, we apply moderate batch mixing: each batch is paired with a partner from a nearby stage that has similar average difficulty score and sequence length but a different dominant cluster, and the last $h$ samples of the two batches are exchanged ($h\ll b$). The mixing strength $h$ trades off focus on a single region against coverage of other regions; we find that moderate mixing performs best (\Cref{subsec:batch_diversity_exp}), consistent with a bias--variance view (Appendix~\ref{app:sae_batch_tradeoff}). Further details are provided in Appendix~\ref{app:curriculum-mixing}.

\section{Main Experiment}
We evaluate \method in the mathematical reasoning domain from two perspectives: downstream performance, compared with data selection and curriculum learning baselines, and training efficiency, measured by the number of training steps needed to reach a target accuracy. We also examine whether the curriculum constructed with small models transfers to a larger model.

\subsection{Experiment Setup}
\label{sec:setup}

\paragraph{Models and training.}
We train two model scales, \texttt{Qwen2.5-Math-1.5B} and \texttt{Qwen2.5-Math-\allowbreak7B}~\citep{yang2024qwen25math}, on \texttt{DeepMath-103K}~\citep{he2025deepmath} with a batch size of $128$ to test the generality of \method.
We use GRPO~\citep{shao2024deepseekmath} as the default RL algorithm.
We denote the selection-only variant as \methodsel, which trains on the selected subset in random order; the curriculum-only variant as \methodcur, which applies both ordering and batching to the full dataset; \methodord and \methodbatch, which apply only difficulty-based ordering or only SAE-space batching to the full dataset, respectively; and the complete framework as \method, which applies the full curriculum to the selected subset.
% We denote \method trained with GRPO ~\citep{shao2024deepseekmath} and DAPO ~\citep{yu2025dapo} as \textbf{\methodg} and \textbf{\methodd}, respectively. We train an SAE on layer-27 activations of \texttt{Qwen3-1.7B} \citep{yang2025qwen3} as the shared encoder for all data engineering operations, demonstrating that a single SAE trained on one model can effectively guide post-training data engineering for other model families and larger scales. 
We train SAEs on layer-27 activations of \texttt{Qwen2.5-Math-1.5B} and \texttt{Qwen3-1.7B}~\citep{yang2024qwen25math,yang2025qwen3} for data selection and curriculum construction, respectively. This setup tests whether SAE representations from small models can organize training data for other model families and larger scales.
Additional details are provided in Appendix~\ref{app:hyperparameters}.

\paragraph{Evaluation.}
We evaluate on six benchmarks spanning a wide difficulty range: \texttt{GSM8K}~\citep{cobbe2021gsm8k} and \texttt{AMC23} (lower), \texttt{MATH500}~\citep{lightman2023lets} and \texttt{MinervaMath}~\citep{lewkowycz2022minerva} (mid), and \texttt{OlympiadBench}~\citep{he2024olympiadbench} and \texttt{AIME24} (competition-level), which are referred to as \texttt{GSM8K}, \texttt{AMC}, \texttt{MATH}, \texttt{MNV}, \texttt{OLPD}, and \texttt{AIME}, respectively. We report \texttt{Pass@8} for \texttt{AIME24} and \texttt{Avg@8} for the remaining five benchmarks.

\paragraph{Baselines.}
We compare \method against seven baselines.
% Vanilla \textbf{GRPO} ~\citep{shao2024deepseekmath} and \textbf{DAPO} ~\citep{yu2025dapo} serve as RL algorithm baselines without curriculum, and we pair \method with both to test whether its benefits are consistent across RL algorithms. 
Vanilla \textbf{GRPO}~\citep{shao2024deepseekmath} serves as an RL algorithm baseline without selection or curriculum.
For data selection, \textbf{LearnAlign}~\citep{li2025learnalign} combines rollout success rates with gradient alignment to select learnable and representative samples, while \textbf{CROPI}~\citep{zhu2025cropi} uses compressed off-policy gradients to estimate each sample's contribution to the current policy update. Both provide gradient-based comparisons. \textbf{IRDS}~\citep{li2026irds} selects samples through verifier-coupled coverage of SAE feature clusters, providing an SAE-based selection comparison.
For curriculum construction, \textbf{Difficulty Curriculum Learning}~\citep{narvekar2020curriculum} orders samples by the difficulty labels provided with the dataset, representing curricula built on external annotations. \textbf{ADARFT}~\citep{shi2025adarft} estimates difficulty from rollout accuracy, representing rollout-based curriculum methods. \textbf{GAINRL}~\citep{wang2025angles} schedules training samples by the angle concentration of their hidden-state representations, representing curricula built on model internals.

\subsection{Training Performance}
\label{subsec:model_performance}
% !TeX root = ../0.main.tex
\begin{table*}[!ht]
\definecolor{saerlavgcolumn}{HTML}{DFEFF8}
\definecolor{saerlavgheader}{HTML}{BFDEF1}
\small
\setlength{\tabcolsep}{2.6pt}
\renewcommand{\arraystretch}{1.05}

\centering
\resizebox{\textwidth}{!}{%
\begin{tabular}{
>{\centering\arraybackslash}m{0.90cm}
>{\centering\arraybackslash}m{1.55cm}
>{\centering\arraybackslash}m{1.50cm}
rrrrrr>{\columncolor{saerlavgcolumn}}r
@{\hspace{12pt}}
rrrrrr>{\columncolor{saerlavgcolumn}}r
}
\toprule

\multirow[c]{2}{=}{\centering\raisebox{-0.6ex}{Model}}
& \multirow[c]{2}{=}{\centering\raisebox{-0.6ex}{Type}}
& \multirow[c]{2}{=}{\centering\raisebox{-0.6ex}{Method}}
&
\multicolumn{7}{c}{\textbf{Accuracy (\%) $\uparrow$}}
&
\multicolumn{7}{c}{\textbf{Steps to Target $\downarrow$}}
\\

\cmidrule(lr){4-10}
\cmidrule(lr){11-17}

& & & AIME & AMC & GSM8K & MATH & MNV & OLPD
& \cellcolor{saerlavgheader}\textbf{Avg}
& AIME & AMC & GSM8K & MATH & MNV & OLPD
& \cellcolor{saerlavgheader}\textbf{Avg}
\\
\midrule

% =========================================================
% Qwen2.5-Math-1.5B
% =========================================================

\multirow[c]{8}{=}{
    \centering
    \rotatebox[origin=c]{90}{
        \shortstack[c]{Qwen2.5\\Math-1.5B}
    }
}
& -
& GRPO
& 30.0 & 53.4 & 81.1 & 71.4 & 26.1 & 34.8 & 49.5
& \underline{40} & \underline{340} & 560 & 700 & 780 & 800 & 537
\\

\cmidrule(lr){2-17}

& \multirow[c]{3}{=}{\centering Selection}
& LearnAlign
& \textbf{36.7} & 52.2 & 80.5 & 70.7 & 25.0 & 34.6 & 49.9
& \textbf{20} & 420 & 520 & 640 & 720 & 780 & 517
\\

& &
CROPI
& \underline{33.3} & \textbf{55.3} & 80.4 & 70.5 & 25.3 & 34.9 & 50.0
& \underline{40} & 520 & 660 & 720 & 800 & 800 & 590
\\

& &
IRDS
& \underline{33.3} & 53.8 & 81.6 & 71.3 & 25.6 & \underline{35.2} & 50.1
& \textbf{20} & 620 & \underline{460} & 560 & \underline{640} & 700 & 500
\\

\cmidrule(lr){2-17}

& \multirow[c]{3}{=}{\centering Curriculum}
& DIFF
& \underline{33.3} & \underline{55.0} & \underline{81.7} & 72.2
& \underline{26.5} & 34.7 & \underline{50.6}
& 60 & \underline{340} & \underline{460} & 600 & 700
& \underline{620} & \underline{463}
\\

& &
ADARFT
& \underline{33.3} & 52.8 & \underline{81.7} & \underline{72.3}
& 25.8 & 34.8 & 50.1
& \underline{40} & 620 & \underline{460} & \underline{540}
& 680 & 800 & 523
\\

& &
GAINRL
& \underline{33.3} & 53.1 & 79.2 & 70.9 & 25.6 & 35.0 & 49.5
& \underline{40} & 600 & 780 & 800 & 780 & 840 & 640
\\

\cmidrule(lr){2-17}

& Joint
& \textbf{\method}
& \textbf{36.7} & \textbf{55.3} & \textbf{85.4}
& \textbf{72.7} & \textbf{29.4} & \textbf{36.0} & \textbf{52.6}
& \textbf{20} & \textbf{200} & \textbf{320} & \textbf{380}
& \textbf{380} & \textbf{560} & \textbf{310}
\\

\midrule

% =========================================================
% Qwen2.5-Math-7B
% =========================================================

\multirow[c]{7}{=}{
    \centering
    \rotatebox[origin=c]{90}{
        \shortstack[c]{Qwen2.5\\Math-7B}
    }
}
& -
& GRPO
& 46.7 & 68.1 & 90.4 & 79.2 & 33.7 & 42.1 & 60.0
& \underline{40} & 480 & 300 & \underline{280} & 500 & 280 & 313
\\

\cmidrule(lr){2-17}

& \multirow[c]{2}{=}{\centering Selection}
& LearnAlign
& 50.0 & 66.6 & \textbf{91.4} & \underline{79.4}
& \textbf{36.1} & \underline{42.3} & 61.0
& 120 & 880 & 320 & 300
& \underline{260} & 300 & 363
\\

& &
CROPI
& \textbf{56.7} & 67.5 & 88.8 & 77.5
& 33.0 & 39.4 & 60.5
& \underline{40} & \underline{380} & 700 & 480
& 780 & 680 & 510
\\

\cmidrule(lr){2-17}

& \multirow[c]{3}{=}{\centering Curriculum}
& DIFF
& 50.0 & \underline{68.8} & \underline{90.9} & 79.1
& 33.0 & \textbf{42.7} & 60.7
& \textbf{20} & 760 & 340 & 300 & 900 & 260 & 430
\\

& &
ADARFT
& \underline{53.3} & 66.9 & 88.9 & 79.0
& 32.8 & 40.9 & 60.3
& \textbf{20} & \textbf{320} & 640 & 300
& 420 & \underline{240} & 323
\\

& &
GAINRL
& \underline{53.3} & 68.4 & 90.1 & \textbf{79.8}
& 34.7 & 41.8 & \underline{61.4}
& 80 & 440 & \underline{260} & \textbf{200}
& \textbf{240} & 280 & \underline{250}
\\

\cmidrule(lr){2-17}

& Joint
& \textbf{\method}
& \underline{53.3} & \textbf{70.6} & 90.3 & \textbf{79.8}
& \underline{34.8} & \underline{42.3} & \textbf{61.9}
& \textbf{20} & 400 & \textbf{240} & \textbf{200}
& \underline{260} & \textbf{220} & \textbf{223}
\\

\bottomrule
\end{tabular}%
}

\caption{
Training performance and efficiency on six mathematical reasoning benchmarks.
Accuracy (\%) is reported at training step 900.
For each model--benchmark pair, the target accuracy is defined as the minimum final accuracy across all compared methods, ensuring that every method reaches the target by step 900.
Methods are categorized into selection, curriculum, and joint data-engineering approaches.
DIFF denotes Difficulty Curriculum Learning.
Higher accuracy and fewer training steps are better.
\textbf{Bold} and \underline{underline} denote the best and second best results, respectively.
}
\label{tab:acc_speedup}

\end{table*}

% \Cref{tab:main_result} shows that \method improves average accuracy across RL algorithms, baselines, and model scales. 
% At the \(1.5\mathrm{B}\) scale, \method improves both GRPO and DAPO, showing that the SAE-based curriculum is not specific to a particular RL algorithm. Compared with Difficulty Curriculum Learning, ADARFT, and GAINRL, \method obtains stronger overall performance, indicating that sparse SAE activations provide a more useful signal than external difficulty labels, rollout accuracy, or compressed hidden states. At the \(7\mathrm{B}\) scale, \methodg again achieves the best average result among the compared methods, suggesting that a shared SAE trained on a smaller model can still guide data engineering for larger models. 

\Cref{tab:acc_speedup} shows that \method consistently improves average accuracy across baseline families and model scales.
At the \(1.5\mathrm{B}\) scale, \method outperforms both gradient-based selectors and the SAE-based IRDS baseline, as well as curricula based on external difficulty labels, rollout accuracy, or hidden-state geometry.
The ablation in \Cref{tab:ablation} further shows that each operation alone already surpasses the corresponding baselines, indicating that organizing data in the SAE feature space benefits both selection and curriculum construction.
At the \(7\mathrm{B}\) scale, \method achieves the highest average accuracy, suggesting that data organized with SAEs from smaller models remains effective for training larger models.
Notably, \method uses only about \(43\%\) of the full dataset yet surpasses full-data GRPO in average accuracy at both model scales, suggesting that SAE-gradient selection retains a compact subset with more useful training signals than the full unfiltered dataset.

\subsection{Training Efficiency}
\label{subsec:training_efficiency}

\method improves training efficiency by reducing both training steps and preparation cost.

% \Cref{tab:speedup} evaluates convergence speed by measuring how many training steps each method needs to reach a shared target accuracy. At the \(1.5\mathrm{B}\) scale, \method accelerates both GRPO and DAPO. \methodd gives the fastest average convergence, and \methodg requires fewer average steps than GRPO, ADARFT, and GAINRL. At the \(7\mathrm{B}\) scale, \methodg reaches the target in the fewest average steps. These results show that SAE-guided data engineering improves convergence across different model scales and RL algorithms.
\Cref{tab:acc_speedup} demonstrates that \method is the most efficient method at both model scales.
% Within each baseline family, \methodsel gives the largest speedup among selection methods at both scales, while \methodcur provides the strongest acceleration among curriculum methods on the smaller model. These results show that SAE signals accelerate training when used for either data selection or curriculum construction.
% Combining both components brings further acceleration, making \method the fastest method on the larger model. Since its SAE signals are derived from a smaller model, this result also demonstrates that efficiency gains transfer to larger models.
At the \(1.5\mathrm{B}\) scale, \method reaches the target accuracy in the fewest training steps on every benchmark, suggesting that SAE-guided data engineering yields consistent convergence gains. At the \(7\mathrm{B}\) scale, it achieves the fastest average convergence, indicating that the curriculum constructed with smaller models also accelerates training on a larger model.

\method also reduces preparation cost. 
\method performs selection once offline and reuses the selected subset throughout RL training, avoiding the repeated gradient computation and data reranking required by CROPI at every curriculum round. Moreover, it constructs the subset using SAE gradients from the \(1.5\mathrm{B}\) model and transfers it directly to \(7\mathrm{B}\) training, avoiding selection-time rollouts and gradient computation on the larger model. 
Regarding curriculum construction, the Difficulty baseline and ADARFT achieve comparable convergence speed but require LLM-generated labels or multiple rollouts per problem at substantial cost. 
In contrast, \method trains the difficulty proxy from a small labeled subset of \(3{,}000\) samples, and SAE encoding for the full dataset of \(103{,}022\) samples takes about \(1\) A800 GPU hour. 
Thus, \method obtains its convergence gains with substantially lower preprocessing overhead.

To test robustness beyond the default setup, we vary the RL algorithm and batch size for \method and the SAE input span, source model, and the difficulty supervision source for \methodcur. Across these settings on \texttt{Qwen2.5-Math-1.5B}, both variants improve average accuracy and convergence over their respective baselines, demonstrating that SAE-guided data engineering remains effective across training configurations, SAE representations, and difficulty supervision sources (Appendix~\ref{app:additional-robustness}).

\section{Analysis}
We analyze the sources of \method's gains from four aspects: component contributions, representation space, batch composition, and interpretability.

\subsection{Ablation Study}

\method combines data selection with a curriculum consisting of ordering and batching.
% Curriculum ordering defines the easy-to-hard trajectory, cluster-first grouping preserves local coherence in SAE activation space, and moderate batch mixing adds limited cross-cluster coverage without disrupting the trajectory.
Data selection retains samples with strong, non-redundant gradient signals, ordering defines the easy-to-hard trajectory, and batching balances within-batch coherence and cross-batch coverage in SAE space.

% !TeX root = ../0.main.tex
\begin{table*}[!ht]
\definecolor{ablationavgcolumn}{HTML}{DFEFF8}
\definecolor{ablationavgheader}{HTML}{BFDEF1}
\small
\setlength{\tabcolsep}{2.6pt}
\renewcommand{\arraystretch}{1.05}

\centering
\resizebox{\textwidth}{!}{%
\begin{tabular}{
>{\raggedright\arraybackslash}m{2.20cm}
rrrrrr>{\columncolor{ablationavgcolumn}}r
@{\hspace{12pt}}
rrrrrr>{\columncolor{ablationavgcolumn}}r
}
\toprule

\multirow[c]{2}{=}{\centering\raisebox{-0.6ex}{Method}}
& \multicolumn{7}{c}{\textbf{Accuracy (\%) $\uparrow$}}
& \multicolumn{7}{c}{\textbf{Steps to Target $\downarrow$}}
\\

\cmidrule(lr){2-8}
\cmidrule(lr){9-15}

& AIME & AMC & GSM8K & MATH & MNV & OLPD
& \cellcolor{ablationavgheader}\textbf{Avg}
& AIME & AMC & GSM8K & MATH & MNV & OLPD
& \cellcolor{ablationavgheader}\textbf{Avg}
\\
\midrule

GRPO
& 30.0 & 53.4 & 81.1 & 71.4 & 26.1 & 34.8 & 49.5
& \underline{40} & 340 & 840 & 800 & 780 & 800 & 600
\\

\midrule

\methodsel
& \underline{36.7} & \underline{55.3} & \underline{85.3}
& \textbf{73.4} & \underline{27.8} & \underline{35.8} & 52.4
& \textbf{20} & 440 & \underline{460} & \underline{520}
& \underline{540} & \underline{580} & \underline{427}
\\

\methodcur
& \textbf{40.0} & \textbf{56.3} & 83.6 & 72.0
& 27.3 & \underline{35.8} & \underline{52.5}
& \textbf{20} & \underline{220} & 520 & 580 & 660 & 620 & 437
\\

\hspace{1em}\methodbatch
& 33.3 & 52.2 & 81.7 & 71.4 & 25.1 & 35.1 & 49.8
& 60 & 340 & 800 & 780 & 760 & 800 & 590
\\

\hspace{1em}\methodord
& \underline{36.7} & 55.0 & 82.3 & 71.3 & 25.3 & 34.5 & 50.9
& 80 & 240 & 840 & 860 & 740 & 900 & 610
\\

\midrule

\textbf{\method}
& \underline{36.7} & \underline{55.3} & \textbf{85.4}
& \underline{72.7} & \textbf{29.4} & \textbf{36.0} & \textbf{52.6}
& \textbf{20} & \textbf{200} & \textbf{360} & \textbf{480}
& \textbf{380} & \textbf{560} & \textbf{333}
\\

\bottomrule
\end{tabular}%
}

\caption{
Ablation results on \texttt{Qwen2.5-Math-1.5B}, reported in accuracy (\%) at step 900 and training steps to target.
\textbf{Bold} and \underline{underline} denote the best and second best results, respectively.
}
\label{tab:ablation}

\end{table*}

% \Cref{tab:ablation} shows that removing difficulty sorting causes the largest degradation, confirming that the easy-to-hard trajectory is central to \method. The \mbox{w/o Clus \& Mix} variant removes cluster assignments and therefore cannot perform moderate batch mixing, leaving a difficulty-only curriculum. Its drop indicates that difficulty sorting alone is insufficient, and SAE-space grouping provides useful local coherence.

% Comparing \mbox{w/o Diff} with \mbox{w/o Diff \& Mix} shows that mixing without difficulty sorting does not improve the curriculum and can even weaken it. 
% In contrast, the full \method outperforms the variants that remove either difficulty sorting or cluster-based batch construction, indicating that moderate batch mixing is most effective when it is applied on top of an already structured cluster-first, easy-to-hard curriculum.
\Cref{tab:ablation} shows that both \methodsel and \methodcur improve accuracy and convergence over vanilla GRPO, confirming that data selection and curriculum construction are each effective on their own. Combining them in \method further reduces the average steps to target from \(427\) and \(437\) to \(333\) while slightly improving accuracy, indicating that the two operations are complementary. Within the curriculum, \methodord and \methodbatch yield limited gains alone (\(50.9\) and \(49.8\)), whereas combining them in \methodcur improves both accuracy and convergence. The two play complementary roles: ordering provides an easy-to-hard progression, whereas batching maintains within-batch coherence and cross-batch coverage at each stage.

\subsection{Representation Space Analysis}

% !TeX root = ../0.main.tex

\begin{table*}[t]
\definecolor{repavgcolumn}{HTML}{DFEFF8}
\definecolor{repavgheader}{HTML}{BFDEF1}

\footnotesize
\setlength{\tabcolsep}{3.5pt}
\renewcommand{\arraystretch}{1.08}
\centering

\resizebox{0.8\textwidth}{!}{%
\begin{tabular}{
>{\raggedright\arraybackslash}m{1.6cm}
>{\raggedright\arraybackslash}m{3.2cm}
rrrrrr
>{\columncolor{repavgcolumn}}r
}
\toprule
Operation & Representation
& AIME & AMC & GSM8K & MATH & MNV & OLPD
& \cellcolor{repavgheader}\textbf{Avg} \\
\midrule

GRPO
& --
& 30.0 & 53.4 & 81.1 & 71.4 & 26.1 & 34.8 & 49.5 \\

\midrule

\multirow{2}{*}{Selection}
& Hidden-state gradient
& \underline{36.7} & 54.4 & 81.9 & 71.5 & 25.6 & 34.5 & 50.8 \\

& SAE gradient
& \underline{36.7} & \underline{55.3} & \textbf{85.3} & \textbf{73.4}
& \textbf{27.8} & \textbf{35.8} & \underline{52.4} \\

\midrule

\multirow{2}{*}{Curriculum}
& Hidden-state activation
& 33.3 & 55.0 & 82.2 & 71.4
& 25.5 & 34.2 & 50.3 \\

& SAE activation
& \textbf{40.0} & \textbf{56.3} & \underline{83.6} & \underline{72.0}
& \underline{27.3} & \textbf{35.8} & \textbf{52.5} \\

\bottomrule
\end{tabular}%
}

\caption{
Comparison of hidden-state and SAE representations for data selection and curriculum construction on
\texttt{Qwen2.5-Math-1.5B} at step 900.
Selection compares hidden-state gradients with SAE gradients, and curriculum construction compares hidden-state activations with SAE activations.
\textbf{Bold} and \underline{underline} denote the best and second-best results in each benchmark, respectively.
}
\label{tab:sae-vs-dense}

\end{table*}

To test whether SAE decomposition provides more useful data engineering signals than the underlying hidden states, \Cref{tab:sae-vs-dense} compares hidden-state and SAE gradients for selection, and hidden-state and SAE activations for curriculum construction. Both types of signals improve over GRPO, but the SAE variants achieve higher average accuracy in each operation. By decomposing dense hidden states into sparse features, SAEs can mitigate superposition and expose clearer feature-level signals~\citep{elhage2022superposition}. This feature-level structure makes gradient signals more discriminative for selecting valuable samples and activation patterns more informative for curriculum construction, accounting for the gains over dense hidden-state representations.

\subsection{Batch Diversity Analysis}
\label{subsec:batch_diversity_exp}

\begin{figure}[tp]
    \centering
    \includegraphics[width=0.72\linewidth]{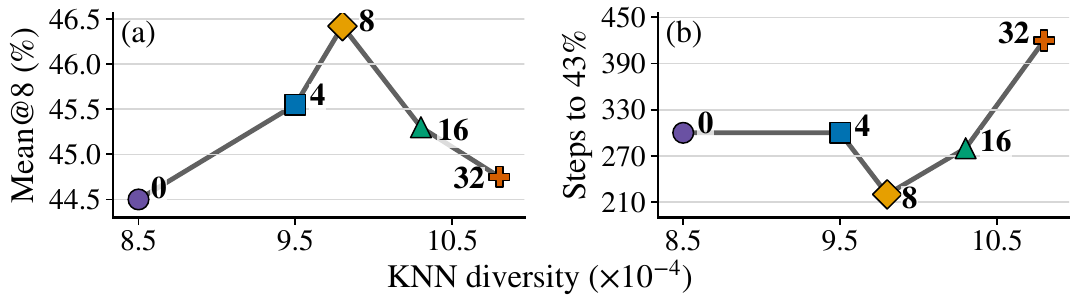}
    \caption{SAE-space batch diversity versus downstream reinforcement learning performance.
    \textbf{(a)} Average \texttt{Avg@8} at step \(800\) as a function of the mean in-batch \(k\)-NN distance in SAE space, with \(k=5\).
    \textbf{(b)} Number of training steps required to reach the fixed average \texttt{Avg@8} threshold \(\alpha=43.0\%\).
    Moderate diversity, represented by \texttt{mix8}, achieves the best step-\(800\) performance and the fastest threshold crossing.}
    \label{fig:batch_diversity}
\end{figure}

The cluster-first curriculum introduces moderate cross-cluster batch mixing to balance within-batch gradient coherence and cross-cluster coverage. The mixing strength is controlled by the number of tail samples swapped between batches. To examine how mixing strength affects downstream performance, we compare five curriculum variants that differ only in this parameter (\texttt{mix0}, \texttt{mix4}, \texttt{mix8}, \texttt{mix16}, and \texttt{mix32}), where \texttt{mix0} is the cluster-first curriculum with no mixing, and larger indices correspond to stronger cross-cluster mixing. All other components of \method are held fixed.

We quantify batch diversity by the mean in-batch $k$-NN distance ($k{=}5$) computed in the two-dimensional SAE projection space, and measure downstream performance by the average \texttt{Avg@8} across the six evaluation benchmarks used in the main experiments. Figure~\ref{fig:batch_diversity} reports both peak performance at step 800 and the number of steps required to reach a fixed threshold $\alpha = 43.0\%$.

The results reveal a clear non-monotonic relationship. Performance improves steadily from \texttt{mix0} to \texttt{mix8}, and \texttt{mix8} reaches $\alpha$ in the fewest training steps. Beyond this point, further increasing the mixing strength to \texttt{mix16} and \texttt{mix32} degrades both step-800 accuracy and convergence speed---despite \texttt{mix32} achieving the highest measured diversity. These results favor moderate mixing among the evaluated configurations and show that increasing the measured diversity alone does not ensure better training outcomes.

A bias--variance analysis (Appendix~\ref{app:sae_batch_tradeoff}) suggests that partial mixing can reduce estimation error, while excessive mixing biases the update away from the reference gradient.

\subsection{Interpretability Analysis}
\label{subsec:interpretability}
\begin{figure}[t]
    \centering
    \includegraphics[width=\linewidth]{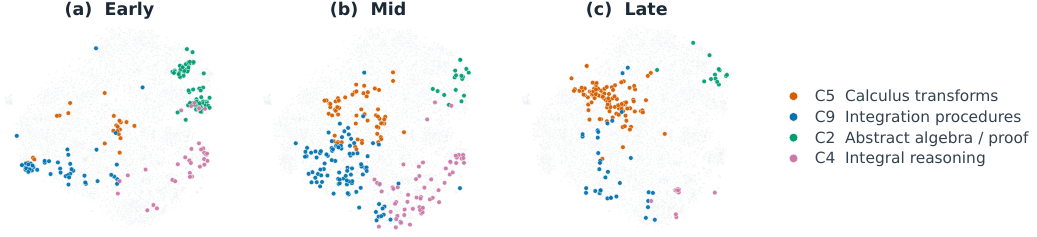}
    \caption{Curriculum order in SAE space. Examples from the \textbf{(a)} early (batches 1--32), \textbf{(b)} middle (batches 159--190), and \textbf{(c)} late (batches 317--348) windows of the curriculum, with four representative clusters highlighted in color; gray points show the fixed background sample. Cluster names are abbreviated from the semantic audit in \Cref{tab:cluster_semantic_summary} and are conservative summaries rather than one-to-one topic labels.}
    \label{fig:curriculum-map-analysis}
\end{figure}

Beyond aggregate performance, we examine how \method organizes data in the SAE feature space over the course of its curriculum. \Cref{fig:curriculum-map-analysis} visualizes three 32-batch windows from the early, middle, and late stages of the curriculum on a shared t-SNE map of SAE representations, with protocol and cluster-audit details in Appendix~\ref{app:interpretability}.

\paragraph{Shifting cluster composition.}
\Cref{fig:curriculum-map-analysis} highlights representative clusters whose shares shift across stages, computed over the complete windows. C2 (abstract algebra and proof) is most prevalent early (\(15.5\%\)) and drops to \(2.8\%\) in the middle; C9 (integration procedures) and C4 (integral reasoning) peak in the middle (\(23.6\%\) and \(14.4\%\)); and C5 (calculus transforms) rises steadily from \(6.9\%\) to \(25.6\%\). No single cluster dominates any stage, consistent with the interleaved design of \method: each batch focuses on one functional region, while successive batches rotate across regions.

\section{Related Work}

We discuss two trends behind our work: post-training data engineering is becoming more adaptive, and model internals are increasingly used to guide training.

\paragraph{Post-training data engineering.}
Post-training data engineering shapes model behavior through two operations, data selection and curriculum construction~\citep{ouyang2022training,lambert2024tulu3,deepseekai2025deepseekr1}. For data selection, many methods filter prompts by rollout pass rates or difficulty~\citep{li2025limr,chen2025samplecentric,sun2025dots,zheng2025greso,zhao2025ufo}, while others score samples by gradient alignment~\citep{li2025learnalign,yang2026gradalign,zhu2025cropi}. For curriculum construction, most methods order samples from easy to hard by difficulty labels, rollout accuracy, or learning progress~\citep{narvekar2020curriculum,tzannetos2024proximal,shi2025adarft,gao2025promptcurriculum,rajaraman2026autocurriculum}, schedule them by hidden-state geometry~\citep{wang2025angles}, or reweight data distributions during training~\citep{wang2025dump}. Others control batch composition through task ratios~\citep{ramesh2026mtgrpo}, variance-based sampling~\citep{leng2025mmr1}, prompt replay~\citep{baroian2026promptreplay}, or explore--exploit allocation~\citep{liang2026leeps}. 
These methods rely on external feedback or raw model signals, whereas \method makes all three decisions with SAE features that indicate which functional regions each sample engages and how training on it would update them.

\paragraph{Model internals for data engineering.}
Model internals have moved from post-hoc analysis toward guiding training. In pre-training and supervised fine-tuning, losses, gradients, and hidden representations guide data filtering and selection~\citep{li2024superfiltering,lin2024rho,xia2024less,ivison2025large,ma2025task,sam2025analyzing,rathi2026shaping,chen2026neuron}. In RL post-training, raw policy gradients and hidden states likewise guide prompt selection and sampling~\citep{li2025learnalign,zhu2025cropi,wang2025angles,liang2026leeps}. 
SAEs decompose hidden representations into sparse features, many of which correspond to human-interpretable concepts~\citep{bricken2023towards,cunningham2023sparse,templeton2024scaling,gao2024scaling}. More recently, SAE activations have been used to select or synthesize training data~\citep{yang2025diversitysae,li2026lessenough,li2026irds}. \method extends them to ordering and batching, combines SAE activations with SAE-projected gradients, and builds the curriculum once with small models for reuse across scales.

\section{Conclusion}

We present \method, which uses model internals, read through SAE features, as feedback for organizing RL post-training data.
Treating groups of co-activated SAE features as functional regions, \method keeps samples with strong, non-redundant gradient signals, trains each region from easy to hard, and trains one region at a time with moderate mixing.
Experiments on mathematical reasoning show consistent gains in accuracy and efficiency across model scales and RL algorithms, with the curriculum built once on small models and transferred to larger ones.
These results suggest that model internals offer rich, model-intrinsic feedback for post-training data engineering, complementary to external feedback.

Future work will extend \method beyond mathematical reasoning to broader post-training settings, better understand how batch composition across functional regions shapes RL updates, and explore how model internals can further reduce reliance on external supervision.

\section*{AI Use Statement}
We use AI assistants in three ways: (1) to aid and polish the writing of this paper; (2) to support literature retrieval and discovery, such as finding related work, with all cited papers checked by the authors; and (3) to assist research ideation and execution, such as discussing ideas and developing code. All AI-assisted content is reviewed and edited by the authors, who remain responsible for the final scientific claims, experiments, and writing.

\section*{Reproducibility Statement}
We have made efforts to ensure that our results are reproducible.
\Cref{sec:sae-rl-pipeline} describes each stage of \method, and \Cref{app:sae-rl-pipeline} provides the technical details of SAE sample representation, clustering, difficulty calibration, curriculum ordering, and batch mixing.
\Cref{sec:setup} specifies the models, training data, RL algorithm, benchmarks, and evaluation metrics, all of which are publicly available.
\Cref{app:detailed-experiment-setup} reports the SAE training details (\Cref{app:sae-training}), all hyperparameters (\Cref{app:hyperparameters}), baseline implementations (\Cref{app:baseline-implementations}), and preprocessing cost (\Cref{app:cost}).
We will release the code for SAE feature extraction, data selection, curriculum construction, training, and evaluation upon publication.

\bibliography{1.custom}
\bibliographystyle{iclr2027_conference}

\appendix

\section{Technical Details of the \method Pipeline}
\label{app:sae-rl-pipeline}

This appendix provides additional technical details for the offline data-processing steps used in Section~\ref{sec:sae-rl-pipeline}. 

\subsection{SAE Sample Representation}
\label{app:sae-representation}

To construct the activation representation, each sample \(x_i\) is divided into a prompt span \(S_i^p\) and a solution span \(S_i^s\).
Let \(\mathcal F\) denote the retained SAE feature set, and let
\(a_t\in\mathbb R^{|\mathcal F|}\) be the SAE activation vector at token \(t\).
For each span, we summarize token-level SAE activations using mean pooling and max pooling.
The pooled SAE activations \(z_i\) are obtained by concatenating the mean-pooled and max-pooled activations from both the prompt and solution spans:
\begin{equation}
z_i=
\big[
\bar a_{S_i^p};
a_{S_i^p}^{\max};
\bar a_{S_i^s};
a_{S_i^s}^{\max}
\big].
\end{equation}
We further append shallow metadata features \(m_i\), including length statistics, TeX ratio, and digit ratio, and use \(\phi(x_i)=[z_i;m_i]\) as the activation representation for clustering and difficulty estimation. 
In our experiments,
\(z_i\) is 960-dimensional and \(m_i\) is 26-dimensional.

To construct the SAE gradient representation, we generate \(R\) rollouts for
each sample \(x_i\) and denote the gradient at response token
\(t\) of rollout \(r\) by \(\gamma_{irt}\in\mathbb R^H\).
We project these gradients onto unit-normalized SAE decoder directions
\(D\in\mathbb R^{F\times H}\), retain the TopK coordinates of each
projection by magnitude, while preserving their signs, and average them over
response tokens and rollouts:
\begin{equation}
    q_i=\frac{1}{R}\sum_{r=1}^{R}
    \frac{1}{T_{ir}}
    \sum_{t=1}^{T_{ir}}
    \operatorname{TopK}(\gamma_{irt}D^\top).
\end{equation}
Here \(T_{ir}\) is the number of response tokens in rollout \(r\). We use \(R=8\) rollouts per sample, and TopK retains the 128 largest-magnitude coordinates of each projection.
We retain the 256 decoder features with the highest mean squared value across
\(\ell_2\)-normalized \(q_i\), forming a shared feature set \(\mathcal G\).
To compare how gradient magnitude is distributed across these features,
we take the element-wise absolute value. The SAE gradient
representation used for selection is
\begin{equation}
    g_i=\lvert q_i[\mathcal G]\rvert\in\mathbb R_{\geq 0}^{256}.
\end{equation}

\subsection{Clustering and Difficulty Calibration}
\label{app:difficulty-clustering-calibration}

We cluster samples in the feature space using MiniBatchKMeans with \(K=10\).
Each sample is assigned to the nearest cluster centroid based on its activation representation \(\phi(x_i)\).

To estimate sample difficulty, we use a small difficulty-labeled subset
\(\mathcal L\) with \(|\mathcal L|=3{,}000\).
For each labeled sample, the difficulty label is denoted by \(d_i^\star\). 
We train an ElasticNet difficulty proxy \(f_D\) on the labeled subset and use it to produce the raw difficulty prediction \(\hat d_i=f_D(\phi(x_i))\).

Because only a limited number of difficulty labels are available, the raw proxy score is calibrated using a global calibration map fitted on the small labeled subset, together with a shrinkage-based cluster residual. Let \(\kappa\) denote the
global calibration map fitted on \(\mathcal L\). 

For each cluster \(c\), we compute the average residual \(\Delta_c\) between the labeled difficulty \(d_i^\star\) and the globally calibrated prediction \(\kappa(\hat d_i)\)
over labeled samples in that cluster. If a cluster has no labeled samples, we set \(\Delta_c=0\). The residual is then scaled by a shrinkage weight
\(\lambda_c=n_c/(n_c+\tau_{\mathrm{sh}})\), where \(n_c\) is the number of labeled samples in cluster \(c\) and \(\tau_{\mathrm{sh}}>0\) controls the strength of shrinkage toward the global calibration. 

The final difficulty score is
\begin{equation}
s_i
=
\kappa(\hat d_i)
+
\lambda_{c_i}\Delta_{c_i}.
\end{equation}
A larger \(s_i\) corresponds to a higher estimated difficulty.

\subsection{Curriculum Ordering and Moderate Batch Mixing}
\label{app:curriculum-mixing}

Within each cluster, samples are sorted by the calibrated difficulty score
\(s_i\) in ascending order, resulting in a local easy-to-hard curriculum. The ordered samples in each cluster are then partitioned into fixed-size batches.
The global curriculum interleaves these batches across clusters stage by stage, which preserves local easy-to-hard trajectories while maintaining coverage across different clusters.

After the cluster-first curriculum has been constructed, we apply moderate batch mixing. For an ordered batch of size \(b\), we keep the first \(b-h\) samples fixed and exchange only the last \(h\) tail samples with another batch.
The partner batch is selected from a local curriculum window and must satisfy three conditions: it should have similar average calibrated difficulty, similar average sequence length, and a different dominant cluster. 
The dominant cluster of a batch is defined as the most frequent cluster label among its samples.

Since only the tail block is exchanged, this operation introduces limited cross-cluster mixing while largely preserving the local curriculum structure within each batch.
In our experiments, the curriculum uses 16-batch difficulty windows and \(h=8\), corresponding to \texttt{mix8} in \Cref{subsec:batch_diversity_exp}.

% \subsection{SAE-based Data Selection}
% \label{app:sae-probe-selection}

% Raw data selection is formulated as a binary classification problem over SAE representations. 
% Let \(\mathcal D_{\mathrm{raw}}\) denote the raw candidate
% pool. 
% For each candidate sample, \(y_i=1\) indicates membership in the target DeepMath-like distribution, while \(y_i=0\) indicates otherwise.

% The quality probe uses only the SAE representation \(z_i\), without metadata.
% We train a linear classifier with SGD and use its predicted probability as the
% selection score:
% \begin{equation}
% s_i
% =
% p_\psi(y_i=1\mid z_i)
% =
% \sigma(w^\top z_i+b),
% \end{equation}
% where \(\psi=(w,b)\) and \(\sigma\) is the logistic sigmoid.

% Given the score \(s_i\), we use either threshold-based selection or fixed-size top-\(k\) selection. The threshold rule selects samples with scores above a quality threshold \(\gamma\), which controls selection precision. The top-\(k\) rule selects the \(k\) highest-scoring samples, which fixes the number of selected samples.

\section{A Bias–Variance View of Moderate Batch Mixing}
\label{app:sae_batch_tradeoff}

We provide a local bias--variance perspective on the potential benefit of moderate cross-cluster batch mixing. The analysis illustrates when partial mixing can improve gradient estimation under a simplified sampling model.

\paragraph{Setup.}
Fix the model parameters $\theta$ and denote the per-sample training gradient by $\tilde g_i=\nabla_\theta\ell(x_i;\theta)$. As a modeling assumption, suppose $\tilde g_i=F(\phi(x_i))+\varepsilon_i$, where $\phi(x_i)$ is the activation representation used for clustering, $F$ is locally Lipschitz, and $\varepsilon_i$ captures variation not explained by $\phi(x_i)$. When residual differences are small, proximity in SAE space can therefore correspond to similar gradients, motivating cluster-based batching. Let $\mu_c$ and $\Sigma_c$ be the gradient mean and covariance within cluster $c$, with finite second moments. We idealize sampling as independent within and across clusters, and let $G_t$ be a fixed reference gradient for the data distribution of interest at the current curriculum stage.

\paragraph{Pure-cluster bias.}
For a batch of size $b$ drawn entirely from cluster $c$, the mean squared error of $\hat G_c=b^{-1}\sum_{i=1}^{b}\tilde g_i$ relative to $G_t$ decomposes as
\begin{equation}
    \operatorname{MSE}_0
    = \|G_t - \mu_c\|^2
    + \frac{1}{b}\operatorname{tr}(\Sigma_c),
    \label{eq:mse0}
\end{equation}
where the two terms are squared cluster-local bias and estimation variance, respectively. A cluster with small $\operatorname{tr}(\Sigma_c)$ yields a low-variance estimate, but its mean may still deviate from the reference gradient.

\paragraph{Effect of mixing.}
Replace $h\in\{0,\ldots,b\}$ samples with independent draws from cluster $d$ and write $\rho=h/b$. With these fixed sample counts, the mixed estimator has
\begin{equation}
    \mathbb E[\hat G_\rho]=(1-\rho)\mu_c+\rho\mu_d,
    \qquad
    \operatorname{Cov}(\hat G_\rho)
    =\frac{(1-\rho)\Sigma_c+\rho\Sigma_d}{b}.
\end{equation}
Let $r_c=G_t-\mu_c$ and $v=\mu_d-\mu_c$. Substituting these moments into the bias--variance decomposition gives the MSE change relative to~\eqref{eq:mse0}:
\begin{equation}
    \Delta(\rho)
    = -2\rho\langle r_c, v\rangle
    + \rho^2 \|v\|^2
    + \frac{\rho}{b}\operatorname{tr}(\Sigma_d - \Sigma_c),
    \label{eq:delta}
\end{equation}
\paragraph{When partial mixing helps.}
Write $\Delta(\rho)=-A\rho+C\rho^2$, where
\begin{equation}
    A=2\langle r_c,v\rangle-\frac{1}{b}\operatorname{tr}(\Sigma_d-\Sigma_c),
    \qquad C=\|v\|^2.
\end{equation}
For $C>0$, the continuous relaxation is minimized at $\rho^\dagger=\operatorname{clip}(A/(2C),0,1)$; feasible mixing counts are obtained by evaluating $\rho=h/b$. An interior minimum exists when $0<A<2C$, and mixing improves on the pure-cluster estimate whenever $0<\rho<A/C$ within the feasible range. If $C=0$, the change is linear and a boundary point is optimal, with all choices equivalent when $A=0$.

The alignment term $\langle r_c,v\rangle$ measures whether the partner cluster moves the expected update toward $G_t$, while the covariance difference accounts for the change in sampling variance. Favorable alignment can make a small amount of mixing beneficial. Beyond the optimal balance, further mixing increases the combined error through added variance or excessive displacement of the mean.

\paragraph{Connection to the curriculum.}
This local argument motivates introducing cross-cluster information while retaining most of a batch's original composition. The implementation exchanges difficulty-ordered tails between paired batches rather than drawing independent samples, and subsequent updates change both gradients and optimizer state. The analysis therefore provides a heuristic explanation for the empirical benefit of moderate mixing by clarifying the balance between directional bias correction and changes in estimation variance.

\section{Implementation Details}
\label{app:detailed-experiment-setup}
This appendix provides the training hyperparameters and baseline implementation details used in the experiments.

\subsection{Sparse Autoencoder Training Details}
\label{app:sae-training}

We train the SAE using the OpenSAE framework on layer-$27$ 
activations of \texttt{Qwen3-1.7B}, with an expansion factor 
of $64$ to support fine-grained feature extraction. The training 
corpus consists of \texttt{FineWeb-Edu}~\citep{penedo2024fineweb} and \texttt{Wikipedia}, 
totaling $80$ GB. Training takes approximately $29$ hours on $4$ A100 GPUs, corresponding to $116$ A100 GPU hours.

We also train an SAE on layer-$27$ activations of \texttt{Qwen2.5-Math-1.5B}, with an expansion factor of $64$, using the same training corpus. Training takes approximately $44$ hours on $8$ A100 GPUs, corresponding to $352$ A100 GPU hours.

\subsection{Hyperparameters}
\label{app:hyperparameters}
We train GRPO on both the \texttt{Qwen2.5-Math-1.5B} and \texttt{Qwen2.5-Math-7B}, and DAPO only on \texttt{Qwen2.5-Math-1.5B}. \Cref{tab:hyperparameters} lists the core training hyperparameters.

\begin{table}[t]
\centering
\small
\setlength{\tabcolsep}{4pt}
\renewcommand{\arraystretch}{1.08}
\begin{tabular}{>{\raggedright\arraybackslash}p{0.40\linewidth}>{\raggedright\arraybackslash}p{0.21\linewidth}>{\raggedright\arraybackslash}p{0.24\linewidth}}
\toprule
\textbf{Hyperparameter} & \textbf{GRPO} & \textbf{DAPO} \\
\midrule
Framework & \texttt{verl} & \texttt{verl} \\
Rollout engine & vLLM & vLLM \\
Learning rate & \(10^{-6}\) & \(10^{-6}\) \\
Train batch size & 128 & 128 \\
Max prompt length & 1024 & 1024 \\
Max response length & 3072 & 3072 \\
Sampling temperature & 0.6 & 0.6 \\
Rollouts per sample & 8 & 8 \\
Candidate generation batch & 128 & 384 \\
PPO clip lower / upper & 0.2 / 0.2 & 0.2 / 0.28 \\
Actor KL coefficient & 0.001 & 0 \\
\bottomrule
\end{tabular}
\caption{Core training hyperparameters for GRPO and DAPO. DAPO uses dynamic sampling to retain 128 valid prompts from a candidate generation batch of 384.}
\label{tab:hyperparameters}
\end{table}

\subsection{Baseline Implementations}
\label{app:baseline-implementations}
For the Difficulty Curriculum Learning baseline \citep{narvekar2020curriculum}, we sort the \texttt{DeepMath-103K} dataset \citep{he2025deepmath} by its provided difficulty labels in ascending order and sample progressively throughout training.

To ensure a fair comparison, we train the ADARFT baseline \citep{shi2025adarft} using GRPO \citep{shao2024deepseekmath}. Following the original method, we estimate problem difficulty with \texttt{Avg@128} rollout accuracy and use the resulting scores to construct the curriculum.

\subsection{Preprocessing Cost}
\label{app:cost}

\Cref{tab:cost} compares the preprocessing cost of gradient-based data selection at the \(1.5\mathrm{B}\) scale.
\method extracts and projects SAE gradients once, including rollout generation, which takes \(32.6\) A800 GPU hours.
LearnAlign requires \(61.7\) GPU hours, including warm-up training before gradient extraction, while CROPI requires \(277.4\) GPU hours because it recomputes gradients and reranks data at every curriculum round.
At the \(7\mathrm{B}\) scale, \method reuses the subset selected with the \(1.5\mathrm{B}\) model and incurs no additional selection cost, whereas LearnAlign and CROPI must repeat rollouts and gradient computation on the larger model, so their costs grow further with model size.

\begin{table}[h]
\centering
\begin{tabular}{lc}
\toprule
Method & Selection cost (GPU hours) \\
\midrule
LearnAlign & \(61.7\) \\
CROPI & \(277.4\) \\
\method & \(\mathbf{32.6}\) \\
\bottomrule
\end{tabular}
\caption{Data selection cost on \texttt{Qwen2.5-Math-1.5B} in A800 GPU hours.}
\label{tab:cost}
\end{table}

\paragraph{SAE training cost.}
When this work began, no public SAEs were available for the Qwen family, so we trained both SAEs used by \method ourselves (Appendix~\ref{app:sae-training}).
Open-source SAEs trained on Qwen models have since been released by Qwen-Scope~\citep{qwen2026qwenscope}, so future work along this line can build directly on public SAEs.
We therefore treat SAE training as a one-time cost, exclude it from the comparison above, and report it separately in Appendix~\ref{app:sae-training}.

\section{Additional Robustness Experiments}
\label{app:additional-robustness}

We assess robustness to training batch size, RL algorithm, SAE input, SAE source, and the source of difficulty supervision. Unless noted otherwise, accuracy is measured at step 900 using Pass@8 for AIME and Avg@8 for the other benchmarks. Steps to target are computed against the lowest final accuracy within each benchmark comparison. 

\subsection{Training Batch Size}
\label{app:batch-size}

\begin{table*}[t]
    \definecolor{saerlavgcolumn}{HTML}{DFEFF8}
    \definecolor{saerlavgheader}{HTML}{BFDEF1}
    \small
    \setlength{\tabcolsep}{3.0pt}%
    \renewcommand{\arraystretch}{1.05}%
    \centering
    \resizebox{\textwidth}{!}{%
    \begin{tabular}{
        ccl
        rrrrrr>{\columncolor{saerlavgcolumn}}r
        @{\hspace{12pt}}
        rrrrrr>{\columncolor{saerlavgcolumn}}r
    }
        \toprule
        \multirow[c]{2}{*}{Metric}
        & \multirow[c]{2}{*}{\(B\)}
        & \multirow[c]{2}{*}{Method}
        & \multicolumn{7}{c}{\textbf{Accuracy (\%) $\uparrow$}}
        & \multicolumn{7}{c}{\textbf{Steps to Target $\downarrow$}} \\
        \cmidrule(lr){4-10}
        \cmidrule(lr){11-17}
        & & & AIME & AMC & GSM8K & MATH & MNV & OLPD
        & \cellcolor{saerlavgheader}\textbf{Avg}
        & AIME & AMC & GSM8K & MATH & MNV & OLPD
        & \cellcolor{saerlavgheader}\textbf{Avg} \\
        \midrule

        \multirow{4}{*}{\makecell[c]{Avg\\@8}}
        & \multirow{2}{*}{128} & GRPO
        & 13.8 & 53.4 & 81.1 & 71.4 & 26.1 & 34.8 & 46.8
        & \textbf{520} & 680 & 840 & 800 & 780 & 800 & 737 \\
        & & \textbf{\method}
        & \textbf{15.4} & \textbf{55.3} & \textbf{85.4}
        & \textbf{72.7} & \textbf{29.4} & \textbf{36.0} & \textbf{49.0}
        & 660 & \textbf{480} & \textbf{360} & \textbf{500} & \textbf{520} & \textbf{560}
        & \textbf{513} \\
        \cmidrule(lr){2-17}
        & \multirow{2}{*}{512} & GRPO
        & 12.5 & 51.9 & 80.4 & 70.0 & 24.5 & 33.9 & 45.5
        & 120 & \textbf{80} & 290 & 220 & 260 & 260 & 205 \\
        & & \textbf{\method}
        & \textbf{14.6} & \textbf{54.7} & \textbf{82.6}
        & \textbf{71.1} & \textbf{26.7} & \textbf{34.8} & \textbf{47.4}
        & \textbf{10} & \textbf{80} & \textbf{180}
        & \textbf{130} & \textbf{200} & \textbf{230} & \textbf{138} \\

        \midrule

        \multirow{4}{*}{\makecell[c]{Pass\\@8}}
        & \multirow{2}{*}{128} & GRPO
        & 30.0 & \textbf{85.0} & 94.2 & 86.2 & 43.4 & 52.7 & 65.3
        & 40 & 340 & 600 & 360 & \textbf{260} & 840 & 407 \\
        & & \textbf{\method}
        & \textbf{36.7} & \textbf{85.0} & \textbf{95.1}
        & \textbf{87.2} & \textbf{44.1} & \textbf{54.8} & \textbf{67.1}
        & \textbf{20} & \textbf{200} & \textbf{320}
        & \textbf{320} & 400 & \textbf{320} & \textbf{263} \\
        \cmidrule(lr){2-17}
        & \multirow{2}{*}{512} & GRPO
        & \textbf{36.7} & \textbf{85.0} & 93.4 & 84.6 & 40.4 & 52.5 & 65.4
        & 70 & \textbf{210} & 170 & 100 & \textbf{50} & 120 & 120 \\
        & & \textbf{\method}
        & \textbf{36.7} & \textbf{85.0} & \textbf{94.2}
        & \textbf{86.0} & \textbf{42.3} & \textbf{53.2} & \textbf{66.2}
        & \textbf{30} & 280 & \textbf{100}
        & \textbf{70} & 70 & \textbf{110} & \textbf{110} \\
        \bottomrule
    \end{tabular}%
    }
    \caption{Accuracy and training efficiency across batch sizes and evaluation metrics. Results are evaluated on \texttt{Qwen2.5-Math-1.5B} at step 900 for \(B=128\) and step 300 for \(B=512\), where \(B\) denotes the training batch size. \textbf{Bold} indicates the best result for each batch size and metric.}
    \label{tab:diff-bs}
\end{table*}

\Cref{tab:diff-bs} shows that \method improves average accuracy and convergence over GRPO at both batch sizes under both evaluation metrics, indicating that its gains are not tied to a particular batch size or metric.

\subsection{DAPO Training}
\label{app:dapo}

% !TeX root = ../0.main.tex
\begin{table}[!t]
\definecolor{saerlappavgcolumn}{HTML}{DFEFF8}
\definecolor{saerlappavgheader}{HTML}{BFDEF1}
\small
\setlength{\tabcolsep}{2.6pt}
\renewcommand{\arraystretch}{1.05}
\centering
\resizebox{\textwidth}{!}{%
\begin{tabular}{>{\raggedright\arraybackslash}m{2.65cm}rrrrrr>{\columncolor{saerlappavgcolumn}}r@{\hspace{12pt}}rrrrrr>{\columncolor{saerlappavgcolumn}}r}
\toprule
\multirow[c]{2}{*}{Method}
& \multicolumn{7}{c}{\textbf{Accuracy (\%) $\uparrow$}}
& \multicolumn{7}{c}{\textbf{Steps to Target $\downarrow$}} \\
\cmidrule(lr){2-8}
\cmidrule(lr){9-15}
& AIME & AMC & GSM8K & MATH & MNV & OLPD
& \cellcolor{saerlappavgheader}\textbf{Avg}
& AIME & AMC & GSM8K & MATH & MNV & OLPD
& \cellcolor{saerlappavgheader}\textbf{Avg} \\
\midrule
DAPO
& 40.0 & \textbf{55.6} & 82.0 & 70.8 & 26.7 & 34.5 & 51.6
& \textbf{40} & 580 & 580 & 580 & 540 & \textbf{580} & 483 \\
\textbf{\method\unskip\textsuperscript{DAPO}}
& \textbf{43.3} & 54.1 & \textbf{84.6} & \textbf{72.2} & \textbf{28.5} & \textbf{35.3} & \textbf{53.0}
& 100 & \textbf{560} & \textbf{360} & \textbf{500} & \textbf{400} & \textbf{580} & \textbf{417} \\
\bottomrule
\end{tabular}%
}
\caption{Training performance with DAPO on \texttt{Qwen2.5-Math-1.5B}. \methodd denotes the complete \method data engineering pipeline trained with DAPO. \textbf{Bold} marks the best result.}
\label{tab:app-dapo}
\end{table}

As shown in \Cref{tab:app-dapo}, applying \method to DAPO improves average accuracy and reduces mean steps to target, demonstrating that the gains extend beyond GRPO.

\subsection{Prompt-Only SAE Representation}
\label{app:promptonly}

% !TeX root = ../0.main.tex
\begin{table}[!t]
\definecolor{saerlappavgcolumn}{HTML}{DFEFF8}
\definecolor{saerlappavgheader}{HTML}{BFDEF1}
\small
\setlength{\tabcolsep}{2.6pt}
\renewcommand{\arraystretch}{1.05}
\centering
\resizebox{\textwidth}{!}{%
\begin{tabular}{>{\raggedright\arraybackslash}m{3.20cm}rrrrrr>{\columncolor{saerlappavgcolumn}}r@{\hspace{12pt}}rrrrrr>{\columncolor{saerlappavgcolumn}}r}
\toprule
\multirow[c]{2}{*}{Method}
& \multicolumn{7}{c}{\textbf{Accuracy (\%) $\uparrow$}}
& \multicolumn{7}{c}{\textbf{Steps to Target $\downarrow$}} \\
\cmidrule(lr){2-8}
\cmidrule(lr){9-15}
& AIME & AMC & GSM8K & MATH & MNV & OLPD
& \cellcolor{saerlappavgheader}\textbf{Avg}
& AIME & AMC & GSM8K & MATH & MNV & OLPD
& \cellcolor{saerlappavgheader}\textbf{Avg} \\
\midrule
GRPO
& 30.0 & 53.4 & 81.1 & 71.4 & 26.1 & 34.8 & 49.5
& 40 & 680 & 840 & 800 & 780 & 800 & 657 \\
\cmidrule(lr){1-15}
\multicolumn{15}{l}{\methodcur} \\
\hspace{0.8em}$+$~Prompt\&Response
& \textbf{40.0} & \textbf{56.3} & \textbf{83.6} & \textbf{72.0} & \textbf{27.3} & \textbf{35.8} & \textbf{52.5}
& \textbf{20} & \textbf{580} & 520 & \textbf{540} & 840 & \textbf{760} & 543 \\
\hspace{0.8em}$+$~Prompt
& 36.7 & 53.4 & 83.1 & 71.0 & 25.6 & 35.4 & 50.9
& \textbf{20} & 600 & \textbf{500} & 560 & \textbf{620} & 900 & \textbf{533} \\
\bottomrule
\end{tabular}%
}
\caption{Curriculum performance on \texttt{Qwen2.5-Math-1.5B} with SAE features extracted from different token spans. The indented rows use features from both the prompt and response spans or from the prompt span alone, respectively. \textbf{Bold} marks the best result.}
\label{tab:app-promptonly}
\end{table}

\Cref{tab:app-promptonly} shows that constructing the curriculum from SAE features extracted only from the prompt span improves both average accuracy and convergence over GRPO. Incorporating features from the response span further improves final accuracy, indicating that response information adds useful structure for curriculum construction.

\subsection{Alternative SAE Source}
\label{app:llama}

% !TeX root = ../0.main.tex
\begin{table}[!t]
\definecolor{saerlappavgcolumn}{HTML}{DFEFF8}
\definecolor{saerlappavgheader}{HTML}{BFDEF1}
\small
\setlength{\tabcolsep}{2.6pt}
\renewcommand{\arraystretch}{1.05}
\centering
\resizebox{\textwidth}{!}{%
\begin{tabular}{>{\raggedright\arraybackslash}m{2.65cm}rrrrrr>{\columncolor{saerlappavgcolumn}}r@{\hspace{12pt}}rrrrrr>{\columncolor{saerlappavgcolumn}}r}
\toprule
\multirow[c]{2}{*}{Method}
& \multicolumn{7}{c}{\textbf{Accuracy (\%) $\uparrow$}}
& \multicolumn{7}{c}{\textbf{Steps to Target $\downarrow$}} \\
\cmidrule(lr){2-8}
\cmidrule(lr){9-15}
& AIME & AMC & GSM8K & MATH & MNV & OLPD
& \cellcolor{saerlappavgheader}\textbf{Avg}
& AIME & AMC & GSM8K & MATH & MNV & OLPD
& \cellcolor{saerlappavgheader}\textbf{Avg} \\
\midrule
GRPO
& 30.0 & 53.4 & 81.1 & 71.4 & 26.1 & 34.8 & 49.5
& 40 & 680 & 840 & 800 & \textbf{780} & 800 & 657 \\
\cmidrule(lr){1-15}
\multicolumn{15}{l}{\methodcur} \\
\hspace{0.8em}$+$~Qwen3-1.7B
& \textbf{40.0} & \textbf{56.3} & \textbf{83.6} & \textbf{72.0} & \textbf{27.3} & \textbf{35.8} & \textbf{52.5}
& \textbf{20} & 580 & \textbf{520} & \textbf{620} & 840 & \textbf{760} & \textbf{557} \\
\hspace{0.8em}$+$~Llama-3.1-8B
& 33.3 & 53.8 & 82.3 & 71.5 & 26.2 & 35.3 & 50.4
& 80 & \textbf{520} & 680 & 780 & 820 & 900 & 630 \\
\bottomrule
\end{tabular}%
}
\caption{Effect of the SAE source for curriculum construction on \texttt{Qwen2.5-Math-1.5B}. The two indented rows use SAEs trained on \texttt{Qwen3-1.7B} and \texttt{Llama-3.1-8B} activations, respectively. \textbf{Bold} marks the best result.}
\label{tab:app-llama}
\end{table}

\Cref{tab:app-llama} compares SAE sources under the same curriculum. The Qwen3 SAE yields stronger accuracy, as expected from its closer match to the Qwen2.5-Math training model. Nevertheless, the Llama-3.1-8B SAE also improves average accuracy and convergence over GRPO, showing that SAE-guided curriculum construction remains effective across model families.

\subsection{OOD Difficulty Curriculum}
\label{app:ood-difficulty-curriculum}

% !TeX root = ../0.main.tex
\begin{table}[!t]
\definecolor{saerlappavgcolumn}{HTML}{DFEFF8}
\definecolor{saerlappavgheader}{HTML}{BFDEF1}
\small
\setlength{\tabcolsep}{2.6pt}
\renewcommand{\arraystretch}{1.05}
\centering
\resizebox{\textwidth}{!}{%
\begin{tabular}{>{\raggedright\arraybackslash}m{2.65cm}rrrrrr>{\columncolor{saerlappavgcolumn}}r@{\hspace{12pt}}rrrrrr>{\columncolor{saerlappavgcolumn}}r}
\toprule
\multirow[c]{2}{*}{Method}
& \multicolumn{7}{c}{\textbf{Accuracy (\%) $\uparrow$}}
& \multicolumn{7}{c}{\textbf{Steps to Target $\downarrow$}} \\
\cmidrule(lr){2-8}
\cmidrule(lr){9-15}
& AIME & AMC & GSM8K & MATH & MNV & OLPD
& \cellcolor{saerlappavgheader}\textbf{Avg}
& AIME & AMC & GSM8K & MATH & MNV & OLPD
& \cellcolor{saerlappavgheader}\textbf{Avg} \\
\midrule
GRPO
& 30.0 & 53.4 & 81.1 & 71.4 & 26.1 & 34.8 & 49.5
& 40 & 680 & 840 & 800 & \textbf{780} & 800 & 657 \\
\cmidrule(lr){1-15}
\multicolumn{15}{l}{\methodcur} \\
\hspace{0.8em}$+$~OOD
& 36.7 & 53.1 & 81.7 & \textbf{72.0} & 26.1 & 35.5 & 50.9
& 40 & \textbf{540} & 760 & 740 & 900 & \textbf{740} & 620 \\
\hspace{0.8em}$+$~In-domain
& \textbf{40.0} & \textbf{56.3} & \textbf{83.6} & \textbf{72.0} & \textbf{27.3} & \textbf{35.8} & \textbf{52.5}
& \textbf{20} & 580 & \textbf{520} & \textbf{620} & 840 & 760 & \textbf{557} \\
\bottomrule
\end{tabular}%
}
\caption{Training performance and efficiency of out-of-domain difficulty information for curriculum construction on \texttt{Qwen2.5-Math-1.5B}. \textbf{Bold} marks the best result.}
\label{tab:app-ood-difficulty}
\end{table}

We further examine whether the difficulty signal used for curriculum construction can transfer across datasets. We train the difficulty predictor on $3$K labeled samples from \textsc{DeepScaler} and use the resulting difficulty estimates to construct the \methodcur curriculum on \textsc{DeepMath}. The remaining curriculum construction procedure is kept unchanged.

As shown in Table~\ref{tab:app-ood-difficulty}, the OOD difficulty curriculum improves the average accuracy over GRPO, while reducing the average steps to target. Although it does not match the default \methodcur setting, the OOD variant retains improvements in both final performance and training efficiency. These results suggest that the difficulty signal derived from SAE representations exhibits a degree of cross-dataset transferability and can still provide useful guidance for curriculum construction when the predictor is trained on out-of-domain data.
\section{Interpretability Details of SAE-Guided Curriculum Construction}
\label{app:interpretability}

This appendix supplements the interpretability analysis in
Section~\ref{subsec:interpretability}. We document the curriculum-map
protocol, the relation between SAE clusters and external topic annotations,
the feature groups that provide predictive signal for difficulty estimation,
and the intermediate variables exposed by the curriculum construction
procedure.

\begin{table}[t]
\scriptsize
\setlength{\tabcolsep}{3pt}
\renewcommand{\arraystretch}{1.08}
\centering
\begin{tabular}{c p{0.22\textwidth} p{0.30\textwidth} p{0.32\textwidth}}
\toprule
Cluster & Conservative label & Main evidence & Interpretation \\
\midrule
0 &
Derivative-centered calculus &
Derivative applications; medium-high difficulty; moderate length &
Local change, optimization, and derivative-based transformations. \\

1 &
Number-theoretic symbolic reasoning &
Congruences; long solutions; medium difficulty &
Congruence-style reasoning with symbolic manipulation and extended derivations. \\

2 &
Abstract algebra and proof structure &
Group theory; high topic entropy; medium-high difficulty &
Algebraic structures and proof-oriented reasoning beyond a single subfield. \\

3 &
Discrete and combinatorial reasoning &
Combinatorics; highest mean difficulty &
Counting, construction, and combinatorial proof patterns with high difficulty. \\

4 &
Integral and continuous reasoning &
Integral applications; long solutions; high difficulty &
Integration-related problems with continuous reasoning and longer derivations. \\

5 &
Broad calculus and transformations &
Derivative-related topics; high entropy; long solutions &
A heterogeneous continuous-math group involving functions and transformations. \\

6 &
Limits and sequence-style reasoning &
Limits; highest top-topic share; lowest entropy &
The most topic-concentrated cluster, centered on limits and sequences. \\

7 &
High-load limits and analysis &
Limits; higher difficulty than Cluster 6; long solutions &
More complex and heterogeneous variants of limit or analysis-style reasoning. \\

8 &
Short-form elementary algebra &
Simple equations; lowest mean difficulty; shortest solutions; highest entropy &
A mixed low-to-mid difficulty group with short algebraic structure. \\

9 &
Integration and symbolic procedures &
Integration techniques; medium difficulty; moderate length &
Procedural symbolic transformation, especially integration-related reasoning. \\
\bottomrule
\end{tabular}
\caption{Cluster-level semantic audit of SAE activation groups on DeepMath. Summaries are generated by a GPT-5.4-based agent and manually reviewed by the authors. Labels are conservative summaries rather than one-to-one topic annotations.}
\label{tab:cluster_semantic_summary}
\end{table}

\begin{table}[t]
\scriptsize
\setlength{\tabcolsep}{4pt}
\renewcommand{\arraystretch}{1.05}
\centering
\begin{tabular}{lcc}
\toprule
Feature group & Top 20 & Top 100 \\
\midrule
\texttt{sol\_mean}    & 10 & 36 \\
\texttt{prompt\_max} & 5  & 26 \\
\texttt{prompt\_mean}& 2  & 24 \\
\texttt{sol\_max}     & 2  & 11 \\
\texttt{meta}         & 1  & 3  \\
\bottomrule
\end{tabular}
\caption{Feature-group composition of the top gain-ranked features in an auxiliary LightGBM model trained for difficulty prediction. This model is used only for interpretability analysis.}
\label{tab:feature_group_importance}
\end{table}

\begin{table}[t]
\scriptsize
\setlength{\tabcolsep}{3pt}
\renewcommand{\arraystretch}{1.08}
\centering
\begin{tabular}{p{0.24\textwidth} p{0.33\textwidth} p{0.34\textwidth}}
\toprule
High-gain feature & Observed high-activation pattern & Interpretation \\
\midrule
\texttt{sol\_mean/50831} &
Abstract algebra, rings, groups, and proof-heavy problems &
Solution-side activation associated with abstract algebraic structure and proof load. \\
\addlinespace[2pt]

\texttt{sol\_mean/28006} &
Measure, integration, and high-level analysis &
Solution-side pattern related to advanced analysis and integration-style reasoning. \\
\addlinespace[2pt]

\texttt{prompt\_max/16071} &
Graphs, geometry, and Asymptote-style prompts &
Prompt-side activation capturing localized diagrammatic or formatting cues. \\
\addlinespace[2pt]

\texttt{sol\_mean/122476} &
Combinatorics, graphs, and discrete structures &
Solution-side signal associated with discrete reasoning and combinatorial structure. \\
\addlinespace[2pt]

\texttt{sol\_mean/60349} &
Simple equations and algebraic word problems &
Lower- to mid-difficulty algebraic patterns involving short symbolic reasoning. \\
\addlinespace[2pt]

\texttt{sol\_mean/88548} &
Congruences, primes, and numeric expressions &
Solution-side number-theoretic and numeric reasoning patterns. \\
\bottomrule
\end{tabular}
\caption{Semantic audit of individual high-gain features using high-activation samples. Summaries are produced by a GPT-5.4-based agent and manually reviewed by the authors.}
\label{tab:high_activation_audit}
\end{table}

\subsection{Curriculum-Map Protocol}

The shared background map in \Cref{fig:curriculum-map-analysis} uses a fixed
sample of \(10{,}000\) \textsc{DeepMath} examples. We preprocess the pooled SAE
representation without metadata, retain 50 principal components, and apply
t-SNE with perplexity 50 and seed 42. The same coordinates are reused in all
three panels. The projection is used only to display where active examples lie;
we do not interpret island area or pairwise distance as quantitative evidence.

The curriculum order contains the selected \(44{,}510\)-sample subset arranged
using calibrated difficulty scores and cluster assignments. Its construction uses
a batch size of 128, 16-batch difficulty windows, and adjacent tail-8 mixing.
The early, middle, and late views correspond to batches 1--32, 159--190, and
317--348 of the curriculum order. Their complete-window mean
difficulties are \(3.92\), \(5.86\), and \(8.22\), respectively. The fixed
10K visualization sample intersects these windows in 414, 432, and 395 unique
examples; among these intersections, only examples from the four highlighted
clusters (C2, C4, C5, and C9) are shown in color. Cluster proportions reported
in the main text are computed from all positions in each complete window rather
than from the displayed subset.

The curriculum order fixes the sample multiset, difficulty signal,
outer-window membership, and mixing budget before RL optimization begins.

\subsection{Cluster-Level Structure}

We compare the SAE cluster assignments on DeepMath with the dataset-provided
topic labels. The overall cluster--topic alignment is low: cluster-topic purity is
\(0.1095\), NMI is \(0.0881\), and ARI is \(0.0394\). These results indicate
that SAE clusters do not simply reproduce the dataset's topic taxonomy.

We therefore further inspect each cluster using top leaf topics, topic entropy,
difficulty distribution, solution length, and representative samples.
Table~\ref{tab:cluster_semantic_summary} provides a summary of the resulting
cluster structure.

Some clusters exhibit clear topical enrichment, such as limits,
combinatorics, group theory, and integration. At the same time, clusters also
differ in problem format, solution length, symbolic density, proof orientation,
and mean difficulty. These summaries should therefore be understood as
coarse descriptions of activation-space structure, rather than as one-to-one
topic labels.

This distinction is necessary because the external topic taxonomy and the SAE
representation characterize different aspects of a sample. Topic labels
describe the subject category assigned by the dataset, whereas SAE clusters
are formed according to model-internal activation patterns. A cluster may
therefore contain multiple mathematical topics while still preserving
structure relevant to curriculum construction, such as similar solution
profiles, reasoning formats, or difficulty ranges.

\subsection{Feature-Level Signals}

We next examine which input feature groups provide predictive signal for
difficulty estimation. The curriculum pipeline uses prompt-side and
solution-side SAE features, together with a small set of metadata features.
For feature-group analysis, we train an auxiliary LightGBM model on the same
difficulty-labeled subset and rank input columns by gain. Because gradient-boosted trees capture nonlinear feature interactions, this model serves as a probe of which features carry difficulty-relevant information. It is used
only for interpretability analysis; the actual curriculum ordering uses the calibrated difficulty scores from the ElasticNet proxy described in
Section~\ref{sec:sae-rl-pipeline}.

Table~\ref{tab:feature_group_importance} reports the source distribution of
the top-20 and top-100 gain-ranked features.

The top-ranked features come primarily from SAE activations rather than
metadata. Among the top-20 features, 19 are SAE-derived and 1 is metadata;
among the top-100 features, 97 are SAE-derived and 3 are metadata. Within
these features, solution-side mean features form the largest group, which is
consistent with sustained solution-side activations providing correlational
signals for difficulty estimation. Prompt-side max features also appear
frequently, indicating that localized strong activations in the problem
statement provide additional predictive cues.

To complement the feature-group statistics, we further inspect
high-activation samples for several high-gain SAE features.
Table~\ref{tab:high_activation_audit} summarizes the semantic tendencies
observed in these samples.

These high-activation samples exhibit several recurring patterns, including
abstract algebra, measure and integration, discrete combinatorics, geometric
formats, elementary algebra, and number-theoretic expressions. These
descriptions indicate semantic tendencies only. SAE features are not strictly
monosemantic: the same SAE feature may be activated in different contexts,
and the same mathematical pattern may be distributed across multiple SAE
features.

\subsection{Procedure-Level Inspectability}

The final curriculum order is constructed from a sequence of explicit
intermediate variables. For each sample, the pipeline records its SAE
representation, metadata, cluster assignment, raw difficulty prediction,
calibrated difficulty score, batch assignment, and, when moderate mixing is
applied, its tail-swap partner. These variables allow researchers to inspect
how a sample moves from the representation space to its final position in the
curriculum.

\end{document}